\documentclass{article}
\usepackage{kr}
\usepackage{times}
\usepackage{soul}
\usepackage{url}
\usepackage[hidelinks]{hyperref}
\usepackage[utf8]{inputenc}
\usepackage[small]{caption}
\usepackage{graphicx}
\usepackage{amsmath}
\usepackage{amsthm}
\usepackage{booktabs}
\usepackage{algorithm}
\usepackage{listings}
\usepackage{mdframed,lipsum}

\usepackage{amsmath}
\usepackage{amssymb}
\usepackage[citestyle=numeric-comp]{}

\usepackage{float}
\usepackage{dblfloatfix}

\usepackage[normalem]{ulem}

\usepackage{fancyhdr} 
\mdfdefinestyle{MyFrame}{%
    linecolor=black,
    outerlinewidth=2pt,
    roundcorner=20pt,
    innertopmargin=4pt,
    innerbottommargin=4pt,
    innerrightmargin=4pt,
    innerleftmargin=4pt,
        leftmargin = 4pt,
        rightmargin = 4pt
    backgroundcolor=gray}

\DeclareMathOperator*{\argmin}{arg\,min}

\title{Combinatorial Reasoning: Selecting Reasons \\ in Generative AI Pipelines via Combinatorial Optimization}

\author{%
Mert Esencan$^{1}$\and
Tarun Advaith Kumar$^1$\and
Ata Akbari Asanjan$^{2,3,4}$\and\\
P. Aaron Lott$^{2,4}$\and
Masoud Mohseni$^{5}$\and
Can Unlu$^1$\and
Davide Venturelli$^{2,4}$\and
Alan Ho$^{1,6}$ \\
\affiliations
$^1$Icosa Computing Inc., New York, USA\\
$^2$NASA ARC - Quantum Artificial Intelligence Laboratory (QuAIL), Moffett Field, California, USA\\
$^3$NASA ARC - Data Sciences Group, Moffett Field, California, USA\\
$^4$USRA Research Institute for Advanced Computer Science (RIACS),
Moffett Field, California, USA\\
$^5$LSIP, Hewlett Packard Labs, Milpitas, CA, USA\\
$^6$DataStax, Santa Clara, California, USA\\
\emails
\{mert, tarun, can\}@icosacomputing.com\\
\{aakbariasanjan, plott, dventurelli\}@usra.edu\\
alan.h@datastax.com
}

\begin{document}

\maketitle

\begin{abstract}

Recent Large Language Models (LLMs) have demonstrated impressive capabilities at tasks that require human intelligence and are a significant step towards human-like artificial intelligence (AI). Yet the performance of LLMs at reasoning tasks have been subpar and the reasoning capability of LLMs is a matter of significant debate. While it has been shown that the choice of the prompting technique to the LLM can alter its performance on a multitude of tasks, including reasoning, the best performing techniques require human-made prompts with the knowledge of the tasks at hand. We introduce a framework for what we call Combinatorial Reasoning (CR), a fully-automated prompting method, where reasons are sampled from an LLM pipeline and mapped into a Quadratic Unconstrained Binary Optimization (QUBO) problem. The framework investigates whether QUBO solutions can be profitably used to select a useful subset of the reasons to construct a  Chain-of-Thought style prompt. We explore the acceleration of CR with specialized solvers. We also investigate the  performance of simpler zero-shot strategies such as linear majority rule or random selection of reasons. Our preliminary study indicates that coupling a combinatorial solver to generative AI pipelines is an interesting avenue for AI reasoning and elucidates design principles for future CR methods.

\end{abstract}

\section{Introduction}

The advent of auto-regressive architectures, notably modern LLMs \cite{vaswani_attention_2017}, mark a significant development in the pursuit of human-like AI. These models exhibit a profound capacity for generating human-like responses, positioning them as impressive tools for information processing. However, despite their extensive training and large parameter counts, LLMs inherently lack robust mechanisms for deep reasoning and strategic planning \cite{nye2021show,valmeekam_planning_2023}, which are necessary for applications demanding high-level cognitive functions.
Moreover, the same architecture makes LLMs prone to hallucination—defined as generation that is nonsensical or unfaithful to source material \cite{xu2024hallucination}.

One approach to handling these limitations is to provide additional retrieved context to reduce incorrect generations. Techniques such as retrieval augmented generation (RAG) query a vector database to retrieve source material before generating the LLM response \cite{lewis2020retrieval}. This approach is particularly suitable for knowledge intensive tasks but does not generalize well to reasoning-intensive tasks. 

A parallel area of research is improvements to prompt engineering and response decoding. A new method dubbed Chain of Thought (CoT) \cite{wei_chain--thought_2022} concatenates hand-annotated example responses with reasoning to the query to create the prompt. The responses from the LLM mimic the examples and contain a ‘reasoning path’ followed by the answer. Another method developed as a complementary decoding approach is Self-Consistency, with the idea that marginalizing over several reasoning paths provides the best possible response \cite{wang_self-consistency_2022}. However, these approaches rely heavily on human annotations and the same static examples may not be relevant to different queries. 

To address these limitations and inspired by the conjecture that the human brain performs gradient-free optimization for reasoning and decision-making  \cite{lecun2023positionpaper}, we propose that integrating combinatorial optimization strategies within LLM frameworks could advance their reasoning capabilities and make them more adept at handling tasks requiring strategic thought.

Our research proposes the integration of an external reasoning engine that interfaces with existing LLM pipelines to fully automate the creation of CoT style prompts. As the reasoning engine sits outside the LLM black box, our work is not an attempt to change the foundational auto-regressive architecture of LLMs but a proposed tool to analyze and possibly augment their reasoning faculties through automated prompt engineering. By employing combinatorial optimization, the engine generates structured prompts that guide the LLM towards the correct answer. Our work intersects two distinct fields - generative AI and probabilistic combinatorial optimization - to tackle human level reasoning tasks.  We construct a first of a kind LLM pipeline with physics-inspired solvers and benchmark the pipeline across a variety of well known Natural Language Processing (NLP) reasoning benchmarks.

\begin{figure*}
    \captionsetup{width= 0.75\linewidth}
    \includegraphics[width=1\linewidth]{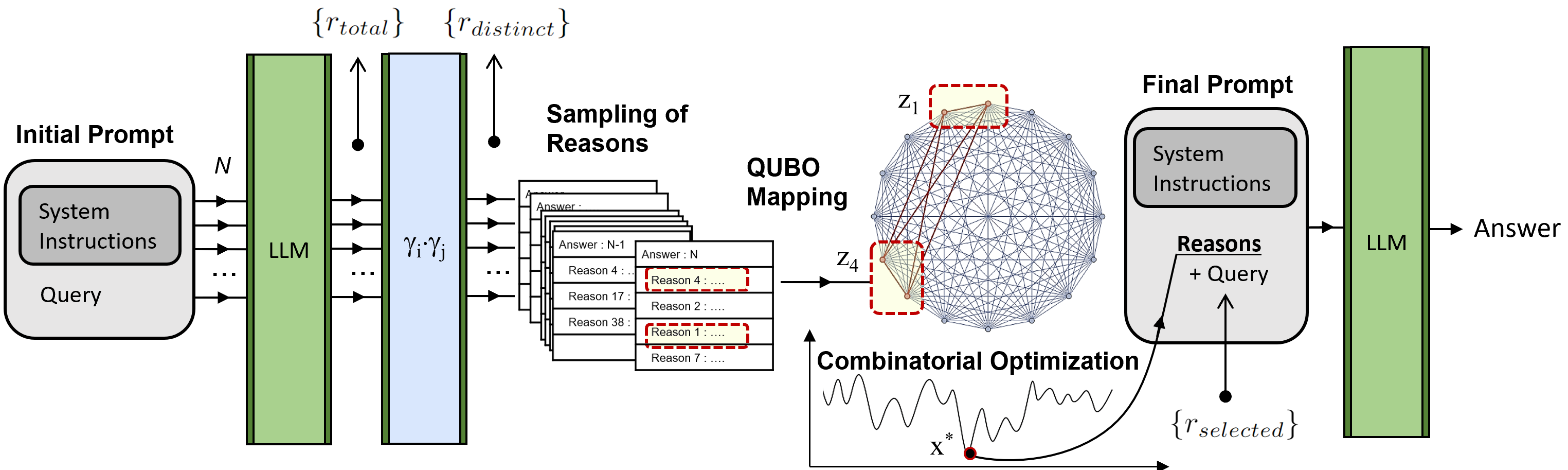}
    \captionof{figure}{Workflow for Combinatorial Reasoning. The initial prompt is processed by the LLM $N$ times and the answers are filtered through a semantic matching procedure to produce answers with distinct reasons. The ensemble is mapped into a QUBO problem solved by an Ising machine. The final solution determines a set of reasons to be added to the prompt for a final LLM call that determines the final answer.}
    \vspace{3mm}
\end{figure*}

In the following sections, we review the state of art and then present our {\bf{Combinatorial Reasoning}} (CR) framework as a versatile technique that leverages a probabilistic combinatorial optimizer to construct a Chain-of-Though style prompt with no human intervention. Our proof-of-concept results demonstrate that in some cases CR achieves improvements over other zero-shot prompting strategies on a few reasoning tasks from BigBench-Hard, and human-level reasoning performance on several reasoning tasks.

In the conclusions, we will discuss the lessons learned, how the CR framework can be further optimized beyond our preliminary baseline experiments and outline multiple promising research avenues. 

\section{Preliminaries and Prior Work}

\subsection{Large Language Models}
Large Language Models (LLMs) are machine learning models that are trained to receive and recognize text as input and produce text as output. Distinguished from simpler language models by their immense parameter count, these models are capable of general purpose language processing tasks. GPT 3.5-turbo, the LLM used in our experiments, is part of a series of models developed by OpenAI for generating human-like natural language text \cite{ye_comprehensive_2023}. Many LLMs including GPT 3.5-turbo are capable of receiving and following a set of system instructions while responding to a query. System instructions differ from the user content that queries an LLM for a response, and we refer to the latter as a `prompt' throughout this work. System instructions define the characteristics of the language model's output, and constrain it to fit requested behavior \cite{GPT4TR}. For our experiments, we use temperature sampling - a popular method for decoding LLMs in text generation tasks. Adjusting the temperature causes the probability distribution of each subsequent token to be modified, affecting the variability of the responses \cite{fan}. High sampling temperatures result in diversity while lower temperatures provide more reproducible generations. 

\subsection{Reasoning in Large Language Models}

There have been many papers that suggest that LLMs can indeed reason \cite{kiciman2023causal,Web2022EmergentReasoning}. For each subsequent revision of LLMs - GPT4 / Gemini / and Llama3, \emph{reasoning} benchmarks such as BIG-Bench-Hard, HellaSwag, and MMLU show ever improving results. However, these results are not a good indicator for the autonomous reasoning capabilities of the model. In each case, the benchmarks are performed using in-context learning, with few-shot (specific examplars) or Chain of Thought (CoT), for which humans manually develop exemplars using labeled datasets to improve performance.

The latest language models do not report the zero-shot performance on these benchmark as in seen Table \ref{tab:gemini} since the performance is likely poorer than those with manual prompts. Thus we believe the next milestone for LLMs is automatic prompt generation with correct reasoning.

\begin{table}[!htp]\centering
\scriptsize
\begin{tabular}{lrrrr}\toprule
&Gemini Ultra &GPT-4 &LLama3 70B \\\cmidrule{2-4}
MMLU &90.04\% CoT@32 &86.4\% 5-shot &79.5\% 5-shot \\\cmidrule{1-4}
GSM8K &94.4\% Maj1@32 &92\% 5-Shot CoT &93.0 8-shot \\\cmidrule{1-4}
MATH &53.2\% 4-shot & &50.4 4-shot \\\midrule
BIG-Bench-Hard &83.6\% 3-shot & &81.3 3-shot, CoT \\
DROP &82.4\% Variable shot &80.9 3-shot &79.7 3-shot,F1 \\
HellaSwag &87.8\% 10-shot &95.3\% 10-shot & \\
WinoGrande & &87.5\% 5-shot &81.3\% 5-shot \\
AI2 Reasoning & &96.3\% 25-shot &93\% 25-shot \\
\bottomrule
\end{tabular}
\caption{Summary of recent reasoning benchmarks on LLMs \protect\cite{gemini_team_gemini:_2023}. Note that reported results are all dependent on some form of In Context Learning or Chain of Thought (CoT) }
\label{tab:gemini}
\end{table}

The main inspiration for our work comes from Yan LeCun's review \cite{lecun2023positionpaper} which suggests multiple models need to work together to emulate general intelligence and that human brain possibly calculates a ``cost function" for reasoning in a gradient-free manner - similar to combinatorial optimization.

\subsubsection{Chain of Thought}
As a natural extension of the in-context few shot learning to facilitate reasoning, the Google Brain team developed few-shot CoT \cite{wei_chain--thought_2022}. The aim is to augment language models with the ability to `reason' in intermediate steps before providing an answer. This approach requires manual demos (or exemplars): labeled question answer pairs containing intermediate reasoning steps that lead to the final correct answer. Inspired by this, \cite{kojima_large_2022} developed ``zero-shot CoT'' as an approach to induce LLMs to answer with intermediate reasoning step without manual demos. Simply put, appending the phrase ``Let's think step by step" to a particular query improved performance. 

\subsubsection{Self-Consistency}

Self-Consistency was introduced by Google's Deepmind Team as an improved decoding approach for CoT style prompts \cite{wang_self-consistency_2022}. Instead of greedy decoding, the authors suggest collecting samples at non-zero temperatures and selecting the most occurring answer. This approach lends itself to an intuitive interpretation - reasoning problems admit multiple correct reasoning paths that lead to the unique right answer but incorrect reasoning paths lead to different incorrect answers. Self-consistency can also be viewed as marginalization over latent tokens to produce a better answer.

\subsubsection{Universal Self-Adaptive Prompting (USP)}
The Universal Self-Adaptive Prompting approach introduces a novel way of generating automatically designed prompts to improve decoding efficiency \cite{wan_universal_2023}. This approach involves the use of unlabeled questions to generate a set of prompt and response pairs. These prompt and response pairs are concatenated to form a collection of pseudo-demos similar in style to the manual demos used in Chain of Thought. For a given question, a selection algorithm picks a subset of this collection - based on computed metrics such as diversity and confidence. These selected demos are then prepended to the beginning of the question in standard CoT style to form the final prompt. 
    
\subsection{Combinatorial Optimization and Ising Machines}

It is well known that challenging combinatorial optimization problems arise in multiple industrial domains, such as finance, logistics, manufacturing, drug design  ~\cite{toshiba_porfolio,venturelli_reverse_2019,comb_opt_trash,protein_folding}. State-of-art solution methods often consist of a patchwork of heuristic techniques tuned to the problem class of interest. Interestingly, for many of these problems efficient mappings of the cost-function $H$ to a binary, quadratic, unconstrained formulation exist (QUBOs~\cite{lucas2014ising,qubo_review}). Equivalently, the problems can be framed in ``physics terms" as approximating as much as possible the ground state of an interacting, disordered classical Ising spin energy function:
\begin{equation}\label{equ:QUBO}
		H = \sum_{\substack{i,j}}Q_{ij} x_i x_j \equiv \sum_{i}h_i s_i + \sum_{\substack{i,j}}J_{ij} s_i s_j,
	\end{equation}
where $x_i\in\{0,1\}$; $s_i\in\{-1,+1\}$ and $J_{ij}$, $h_{i}$, $Q_{ij}$ are real-valued coefficients that specify the problem instance. While the QUBO and Ising forms are equivalent, in this paper we will formulate everything in terms of the QUBO form. Clearly, the search space for the minimum of Eq.~\ref{equ:QUBO} scales as $2^N$ as the number of variables $N$ increases.

The existence of these mappings has ignited a lively research community that, in the last ten years, has devised hardware implementation of the Ising model as well as physics-inspired algorithmic strategies meant to cool down interacting spins close to their least energetic configuration. Collectively, these methods are often referred to as ``Ising Machines''  ~\cite{mohseni2022ising,tanahashi2019ising,tiunov_annealing_2019} or sometimes ``quantum-inspired'' solvers - considering the fact that the most popular and visible methods have a connection to quantum mechanics~\cite{fujitsu_da,nec,hitachi,toshiba_sqbm}. 

Digital implementation of Ising machines are currently the most scalable approach to tackle large problems. This includes GPU/FPGA emulations of solution principles inspired by oscillator synchronization (such as Kuramoto models~\cite{kuramoto}, coherent optics with or without dissipation (e.g Coherent Ising Machines~\cite{{inagaki:cim}} and Bifurcation Machines~\cite{tatsumura2019fpga}) and thermal relaxation (e.g. probabilistic bits (``p-bits'') variations~\cite{camsari_pbits,chowdhury2023full}). Benchmarks of these systems have consistently performed well in paradigmatic benchmarks of combinatorial optimization, especially in absence of hard constraints. Indeed, there is accumulated empirical evidence that on NP-Hard problems such as fully-connected Spin Glasses the time-to-solution scales as a stretched exponential with the increase of the number of binary variables, on typical instances (i.e. $O(N)\propto \exp(\sqrt N)$), while other methods seem to struggle~\cite{sankar2021benchmark,mohseni2022ising}.

\subsubsection{Simulated Annealing and Parallel Tempering}

Simulated Annealing (SA) \cite{sim_anneal83} is an optimization technique built on searching for low energy solutions using a Markov chain parameterized by a temperature such that high temperature samples correspond to random samples and low temperature samples reflect low energy locally-optimal configurations of the target system. A temperature schedule is formulated to effectively explore or exploit regions of the search space for low energy configurations. 
Parallel Tempering (PT) \cite{Mandra_PySA_Fast_Simulated_2023} is a similar optimization technique built on multiple Markov chains running at different temperatures that swap configurations between the chains in order to explore or exploit the temperatures in finding low energy configurations while avoiding stagnation in local minima. 
The hyperparameters for the parallel tempering scheme including the temperature schedule and acceptance rate could also be determined \emph{adaptively} in a fashion to enable a constant swap rate between chains, a scheme dubbed Adaptive Parallel Tempering (APT)~\cite{NMC,Aadit_APT_Adaptive_Parallel_2023}.
The adaptive process is based on a physics-informed procedure to efficiently explore complex energy landscapes. 

\section{Combinatorial Reasoning}

While LLMs cannot reliably reason on their own, with the assistance of an auxiliary system - namely a discrete probabilistic optimizer - we could conceivably select reasons that could create a useful CoT passed into the LLM. The main conceptual challenge is whether one can design a reason-to-variable mapping and a related cost function with the following properties: 

\begin{itemize}
    \item {\bf{universality}}:  works across a large variety of reasoning tasks
    \item {\bf{accuracy}}: its optimized solutions correspond to selecting good reasons when a variety of reasons exist for a given answer
    \item {\bf{practicality}}: its complexity is such that it returns useful reasons within the time allowed for the optimizer to do the minimization
\end{itemize}

With reference to Fig.~\ref{fig:main}, we investigate these challenges by drafting a QUBO cost-function inspired by the problem of portfolio optimization, and designing a sequential procedure of interaction between LLMs and an Ising machine. We call this generic framework Combinatorial Reasoning (CR). It consists of four stages which we now describe in detail.

\subsection{Sampling of Reasons} \label{sec:sampling}

Given a question from the dataset, we prepare $N$ identical input prompts (see appendix \ref{app:sampling}) and query an LLM at a fixed temperature. Following the system instructions, each of the $N$ outputs will contain a set of reasons . Among these, there are duplicate reasons that are semantically equivalent. We use a Sentence Transformer from HuggingFace (\textit{all-mpnet-base-v2}) to embed each reason into a normalized 768 dimensional vector. Defining a similarity metric between two reasons as the dot product of the corresponding embedded vectors, we count two reasons as the same if this metric is greater than $\bf{\zeta}$. Using this procedure as our method for counting, we can reduce the set of all sampled reasons into a smaller set of distinct reasons and a collection, $\{\bf{\gamma}_i\}$, of embedded vectors. We define:

\begin{itemize}
    \item $\{s$\} : Set of samples each with an answer and set of reasons
    \item $\{r_{total}$\} : Set of all reasons sampled from the LLM 
    \item $\{r_{distinct}$\}: Set of independent reasons selected by Sentence Transformer
    \item $n_i$: The number of times each independent reason, indexed by $i$, appears in our $N$ samples
	\item $n_{ij}$: The number of times a pair of independent reasons, indexed by $i$ and $j$, appear together within any one of our $N$ samples
\end{itemize}
These counts are the basis of combinatorial reasoning, and we use these to compute quantities essential in the QUBO mapping. From here on, we refer to independent reasons as reasons for the sake of brevity. Using these counts as well as the acquired embeddings, we denote
$m_i$ as the average similarity that each reason shares with every reason, i.e.
 \begin{equation}
     m_i = \frac{1}{k} \sum_{j = 1}^{k} \bf{\gamma_i} \cdot \bf{\gamma_j}
 \end{equation}
Finally, to clarify our notation, for a given collection $\{\xi_{\mu} \}$, we use $\bar{\xi}$, $\bar{\bar{\xi}}$, and $\delta \xi$ to denote the mean, median, and standard deviation. 

\subsection{QUBO Mapping}\label{qubo_mapping}

This stage processes deterministically the collection of answers and their distinct reasons to formulate a quadratic unconstrained \emph{integer} optimization problem. The procedure that we decide to investigate is inspired by the QUBO mappings to Markowitz portfolio optimization~\cite{grant} where the goal is to select the optimal assets (i.e. reasons, in our case) out of a finite universe (all distinct reasons) maximizing some metric. Importantly, this is just one of the many possible cost-functions that could be designed to try to capture the correlations between good and consistent reasons outputted by an LLM after an ensemble of queries, as it will be discussed in Section 5.
Each distinct reason is associated to an integer variable $z_i$. The integer bound for the variables is a parameter to be chosen as the maximum power of two in order to leverage the binary encoding $z_i=\sum_{w=0}^{W-1}2^w x_{iw}$ where $x_{iw}$ are binary variables.
We now construct two functions that will compose a total objective $H=-(L+Q)$. The first term is meant to select reasons based on their frequency of appearance $n_i/N$:
\begin{equation}
    L =\sum_i l_i(\mu,\alpha)z_i=\sum_{i} \left[\mu\, \mathrm{p}_i - \alpha \mathrm{r}_i\right]z_i
\end{equation}
where $\mathrm{p}_i$ is a measure of ``popularity'', i.e. of the relative deviation with respect to the mean frequency of appearance of a reason. $\mathrm{r}_i$ is a measure of the standard deviation module around the frequency (in analogy to the concept of \emph{risk} in portfolio optimization):
\begin{eqnarray}
\mathrm{p}_i=\frac{n_i-\bar{\bar{n}} }{N}&\;&\mathrm{r}_i^2=\frac{n_i}{N}\left(1-\frac{n_i}{N}\right).\label{eq:p_and_r}
\end{eqnarray}
The real parameters $\mu$ and $\alpha$ needs to be chosen empirically. If this was the only objective function (i.e. $H=-L$) then $z_i$ will either be $0$ or maxed out depending on whether $l_i(\mu,\alpha)$ is positive or negative.

We measure the potential relationship between two reasons by evaluating $c_{ij}$ as the connected correlation between two reasons defined as:
 \begin{equation}
    c_{ij} = \frac{n_{ij}}{N} - \frac{n_i n_j}{N^2}
 \end{equation}

This function assumes values between $-1+N^{-1}$ and $1/4$. It is zero if one of the two reasons appears all the time (meaning it is trivially right - its selection should be handled by linear terms). It is negative when reasons $i$, $j$ are frequent but rarely appearing in the same answer $s$. It hits the maximum positive value when reasons $i$ and $j$ appears all the time together but are not necessarily obvious ($n_i=n_j=N/2$).

Using the connected correlation functions, we construct the second term in $H$ to be sensitive to the correlation between reasons that appear jointly more or less frequently than the average:
\begin{equation}
    Q = \sum_{i \neq j} q_{ij}(\beta)z_i z_j = \sum_{i \neq j} \left[c_{ij} - \bar{c} - \beta\, \delta c\right]z_i z_j\label{eq:L}
\end{equation}
% where the correlation matrix is defined as 
% \begin{equation}
%     q_{ij}(\beta) =  c_{ij} - \bar{c} - \beta\, \delta c .
% \end{equation}

 The parameter $\beta$ is a real valued hyperparameter of the mapping, appearing as a prefactor to the standard deviation of $c_{ij}$. For $\beta=0$, if $c_{ij}$ is positive then $q_{i,j}$ will be positive and increasing with the number of times we observe reasons $i$ and $j$ occur together when this number is greater than the observed average. %\masoud{that is not correct as the $c_ij$ could be negative itself} 
 In a symmetric way, it will be negative and decreasing if $c_{ij}$ is negative and the joint appearance is less frequent than the average. 
 The $L+U$ function is straightforwardly converted to QUBO form Eq.~\ref{equ:QUBO} by means of the binary encoding formula:
 
 \begin{eqnarray}
    H &=& -\sum_i l_i(\mu,\alpha)\sum_{w_1} 2^{w_1} x_{iw_1}\nonumber\\ &-& \sum_{i\neq j} q_{ij}(\beta) \sum_{w_1} \sum_{w_2} 2^{w_1} 2^{w_2} x_{iw_1} x_{jw_2} .\label{eq:Q}
\end{eqnarray}

Despite the simplicity of $L$ and $Q$, it was empirically determined in our study that it is beneficial to use a slightly modified version of the function $\tilde{L}$ introducing flexibility on the mapping of the single non-interacting reasons. Per result of trial and error in our early investigations, the popularity portion was modulated favoring ``crucial'' reasons with low semantic similarity, and the risk portion of the mapping was modified with an additional tuning parameter $\kappa$, as well as other quadratic terms in $x_{iw}$ to modify the integer encodings:

\begin{eqnarray}\label{eq:linear_terms}
    \tilde{L} &=& {\sum_i \tilde{l}_i(\mu,m_i,\alpha,\kappa,x_i)}\nonumber\\ &=& \sum_i \left[ \mu\, \mathrm{sgn}(\mathrm{p}_i)|\mathrm{p}_i|^{m_i}\sum_{w_1}^{W-1} 2^{w_1} x_{iw_1}\right. \\
    &-& \left. \alpha \mathrm{r}_i  \sum_{w_1}^{W-1}\left( 2^{\kappa w_1} x_{iw_1} + \sum_{w_2=w_1+1}^{W-1} 2^{w_1} 2^{w_2} x_{i w_1} x_{i w_2}\right)\right]\nonumber.
\end{eqnarray}

The final QUBO that we send to the solver is then tackling the objective function $\tilde{H}=-(\tilde{L}+Q)$. 

\subsection{Combinatorial Optimization Solver}
The QUBO instance is processed by an Ising machine configured with a pre-defined parameter setting strategy~\cite{neira2024benchmarking} aimed to find the most appropriate solution to the $x_i$ variables. Ideally, the solver identifies the global optimum of $\tilde{H}$, i.e. 
\begin{equation}
    x^\star = \argmin_{\{x_i\}} \left[ \tilde{H}  \right],
\end{equation}
however, an approximate solution might be sufficient as discussed in Section \ref{sec:advanced_annealing}. We denote as $\{r_{selected}$\} the set of reasons selected by QUBO solver. From the binary encoding formula we obtain the $z_i$ variables. We then compose a list including all reasons such that $z_i>0$ in the returned solution, and we associate to each of them a real weight $w$-$value= z_i/Z$ where $Z=\sum_i z_i$.

\subsection{Final Prompt Creation}
Once the best candidate solution to the QUBO problem has been found, it is mapped back to set of reasons $\{r_{selected}$\} each prepended with their $w$-$value$. The LLM is instructed to treat the  $w$-$value$ as a level of relative importance for each reason. These reasons are sorted according to their $w$-$value$s (highest first) and alphabetically. The concatenated string is used to form a prompt (see appendix \ref{app:end2end}). This final prompt inherit the benefits of CoT thanks to these additional reasons, and is used to query the LLM in a zero-shot fashion at temperature = 0 (greedy decoding).

\section{Experimental Results}

\begin{figure*}[t!]
    \centering
    \includegraphics[width = \linewidth]{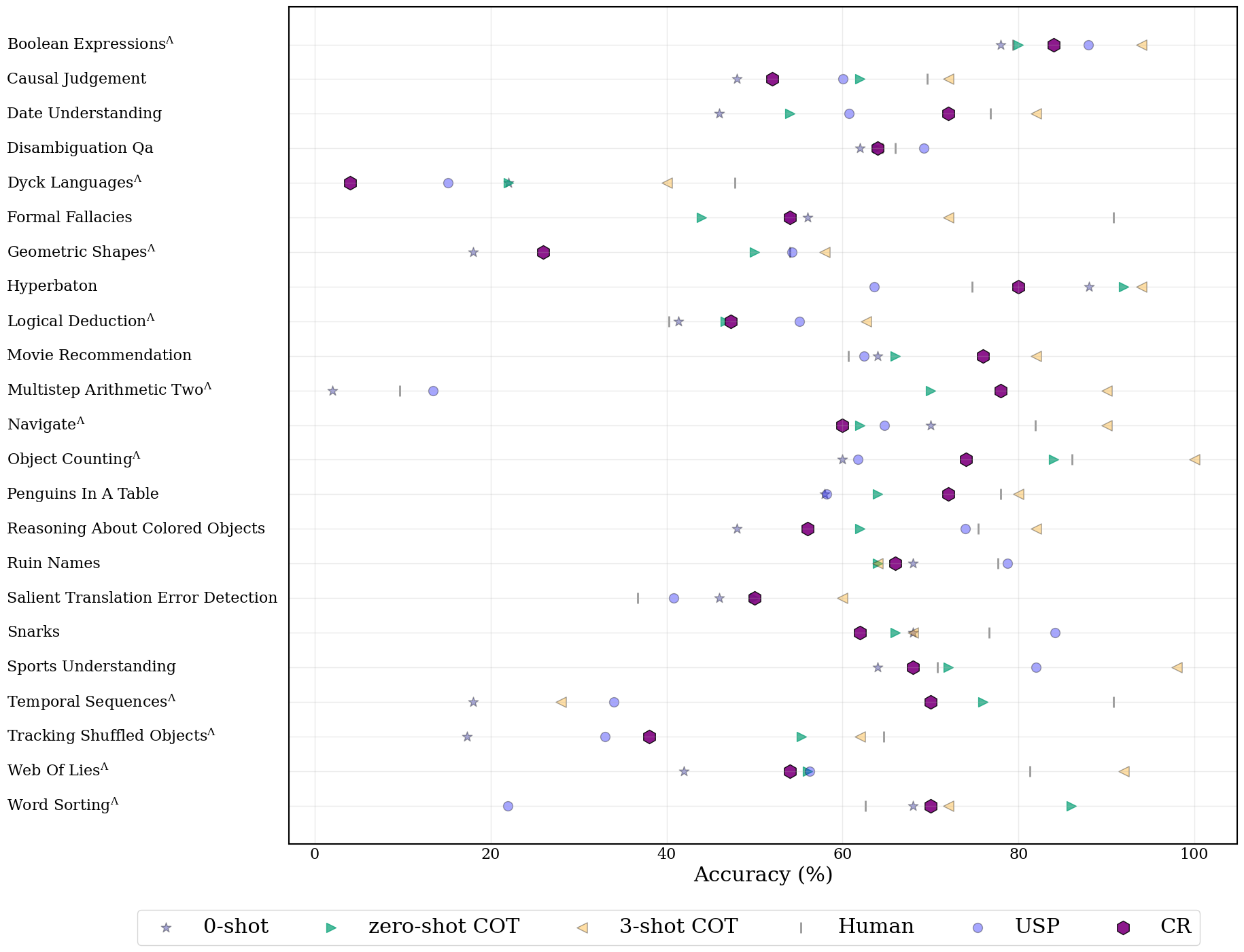}
\captionof{figure}{The performance of combinatorial reasoning (CR) against other methods. Human and USP results are reported from the publications for BBH and USP respectively \protect\cite{wan_universal_2023} \protect\cite{suzgun2022challenging}. USP is evaluated on a different, but comparable, LLM PaLM 2-M. Table \ref{tab:main} presents the cumulative results across BBH for these various tasks. Tasks marked with $\Lambda$ are algorithmic tasks while the others are NLP tasks.}

\label{fig:main}
\end{figure*}

We conduct all of our experiments using the gpt-3.5-turbo-0125 LLM which has a context window of 16,385 tokens and returns a maximum of 4,096 tokens. This language model is a variant of GPT-3.5-Turbo3 produced by OpenAI, and was trained with data available until September 2021.

We selected the suite of BIG-bench Hard (BBH) tasks -  a datasets consisting of reasoning oriented questions that have proven challenging for LLMs in the past \cite{suzgun2022challenging}. To save on inference time and cost, we sample 50 questions from each of the subtasks\footnote{Subtasks Logical Deduction and Tracking Shuffled Objects are split up into three further subtasks, we sample 50 questions from each of these.}, combining them into a 1350 question evaluation set without the subset labels to ensure robustness. On this set, we compare CR against (i) a modified version of zero-shot prompting, (ii) Universal Self-Adaptive Prompting (USP), and (iii) standard three-shot CoT prompting. Our modification to zero-shot consists of an added system-instruction very similar to the one used for CR (see Appendix \ref{app:end2end} for the exact format).

For the Sampling of Reasons step, we sampled the LLM $N = 210$ times at $T = 1$ to collect sufficient distinct reasons, and calculate their distribution and correlations matrices. $N$ was determined empirically on test questions. To map to distinct reason, the similarity threshold is held to $\bf{\zeta}$=0.90, again determined empirically. Prior to running the QUBO mapper, we tune the mapping parameters $\mu$, $\alpha$, $\beta$, $W$ and ($\kappa$ is fixed) using 5 questions from across all of BBH to form a 135 question tuning set. On this, we set the ranges for the tuning (see Table \ref{tab:QUBOparams}) and use \emph{Optuna} - a gradient free hyperparameter optimization framework~\cite{akiba2019optuna} - to select the optimal values for the other four parameters. We note that none of the 135 questions in the tuning set appear in the 1350 question evaluation set. 

\begin{table}[h!]
    \centering
    \resizebox{\linewidth}{!}{%
    \begin{tabular}{c|c|c|c|c|c}
    \toprule
    \textbf{Parameter} & $\mu$ & $\alpha$ & $\beta$ & $W$ & $\kappa$ \\ [0.5 ex]
    \midrule
    \textbf{Tuning Range} & [1E-3, 70] & [1E-4, 20] & [-2, 10] & [1, 4] & 2\\
    \bottomrule
    \end{tabular}}
    \caption{Parameter Tuning Ranges}
    \label{tab:QUBOparams}
\end{table}

For the Ising solver, we utilized an open-source implementation of simulated annealing~\cite{dwave_neal} featuring default settings on temperature, linear annealing schedule, and a fixed parameter setting strategy employing 1000 sweeps, run identically 100 times. 

\begin{table}[!t]
\resizebox{\linewidth}{!}{%
\begin{tabular}{l|rcr|r}
\toprule
\textbf{Setting} & &\textbf{Zero-Shot}&& \textbf{Few-Shot} \\ 
\midrule
Method & 0-Shot & USP & CR & 3-Shot \\
&&&(Ours)&CoT\\
\midrule
Average (\%) $\uparrow$ & 47.68& 55.89& 59.88&74.20\\
Gain over zero-shot  &0&+8.21&+12.20&+26.52\\
Average rank $\downarrow$& 3.22& 2.78& 2.57& 1.35\\
\bottomrule
\end{tabular}}
\caption{Cumulative statistics on the performance of different methods across BBH. CR outperforms the other zero-shot methods both by rank and on average.}
\label{tab:main}
\end{table}
Figure \ref{fig:main} and Table \ref{tab:main} displays our results for BBH tasks. We manually evaluated the results for CR and zero-shot. The USP results are taken from \cite{wan_universal_2023}. While USP was evaluated on PaLM 2-M, we report it here anyway due to its recreation complexity and the superior performance of PaLM 2-M to GPT 3.5 Turbo \cite{gemini_team_gemini:_2023,Palm2}.

We performed a human evaluation at each stage of the CR pipeline. In Table \ref{tab:quadratic percentages} we report the number of sampled reasons before and after the stages depicted in Fig. \ref{fig:main}. It should be noted that the effect of optimization is visible as the mechanism that reduces the number of distinct reasons to a subset of reasons. More results of the human evaluation can be found in the Appendix.

\begin{table}[!b]
\centering
\resizebox{\linewidth}{!}{%
\begin{tabular}{l|c|c|c}
\toprule
 & All Reasons &   & \% of $\{r_{distinct}\}$\\

Dataset & $\{r_{total}\}$& $\{r_{distinct}\}$  & $\{r_{selected}\}$ \\
\midrule
Causal Judgement & 709 & 204 & 87.2 \\
Reasoning About Colored Objects & 525 & 100 & 82.0 \\
Navigate & 1100 & 572 & 100.0 \\
Penguins In A Table & 589 & 123 & 77.2 \\
Geometric Shapes & 630 & 331 & 100.0 \\
Disambiguation QA & 373 & 45 & 68.9 \\
Tracking Shuffled Objects Five Objects & 1020 & 298 & 95.0 \\
Word Sorting & 385 & 107 & 99.1 \\
Tracking Shuffled Objects Three Objects & 743 & 147 & 64.6 \\
Tracking Shuffled Objects Seven Objects & 1164 & 400 & 98.5 \\
Multistep Arithmetic Two & 621 & 253 & 99.6 \\
Web Of Lies & 885 & 113 & 84.1 \\
Logical Deduction Three Objects & 540 & 100 & 72.0 \\
Sports Understanding & 449 & 160 & 96.3 \\
Snarks & 396 & 109 & 91.7 \\
Logical Deduction Five Objects & 680 & 199 & 92.0 \\
Salient Translation Error Detection & 389 & 90 & 98.9 \\
Hyperbaton & 432 & 57 & 65.0 \\
Movie Recommendation & 730 & 457 & 100.0 \\
Object Counting & 397 & 48 & 62.5 \\
Logical Deduction Seven Objects & 730 & 309 & 100.0 \\
Temporal Sequences & 533 & 76 & 97.3 \\
Formal Fallacies & 579 & 251 & 100.0 \\
Dyck Languages & 1112 & 558 & 100.0 \\
Date Understanding & 587 & 162 & 98.1 \\
Boolean Expressions & 493 & 160 & 93.8 \\
Ruin Names & 622 & 421 & 100.0\\
\bottomrule
\end{tabular}}
\caption{Reason filtering and selection percentages}
\label{tab:quadratic percentages}
\end{table}

\section{Conclusion} We propose CR as a zero-shot automatic prompting pipeline applicable to reasoning tasks. We believe CR could be advantageous in the scenario that one needs multiple reasons to elicit the correct answer, and the reasons cannot be obtained via a single-shot from the LLM.

\section{Future Work}

In this section, we point out some details on the upcoming research directions to improve CR.

\subsection{Improving time and accuracy}\label{sec:advanced_annealing}

We list the following straightforward improvement ideas to the baseline framework:

\subsubsection{Semantic matching}
Human evaluation on a few samples ($N=10$) on the \emph{causal judgement}, \emph{movie recommendation}, and \emph{sports understanding} datasets reveals that a decent fraction of the reasons that are identified as distinct via our automated procedure are actually semantically the same in the eyes of a human. This speaks to the relative simpleness of BBH for LLM reasoning and clearly negatively affects the effectiveness to the QUBO mapping and of the entire CR pipeline. It is therefore a priority to improve the semantic matching procedure via threashold adjustment or other more sophisticated filtering.

\subsubsection{QUBO Mapping}

First and foremost, we re-emphasize the fact that the QUBO  construction is just a first attempt to identifying a good objective function. The $\tilde H$ construction can be refined and studied carefully to maximize the correlation between the quality of the $x^\star$ and the accuracy of the final answer at the end of the CR procedure.  Approximate solutions of these hard problems might end up being good enough and optimal with respect to the end-to-end result. In Appendix \ref{append:QUBOs} we study some properties of the current cost-function choice, learning valuable lessons for the design of future improved CR pipelines. It will be also clearly beneficial to study the property of the graphs (size, weight distribution) and correlate characteristics from the physics of spin-glasses (such as the presence of a phase transition~\cite{rieffel2014parametrized,angelini2023limits}) with the final answer. 

It is immediately notable that for a few task categories (Table 4) our the QUBO mapping does not result in a sub-selection of reasons because all distinct reasons are selected to be part of the final prompt. This indicates that either the problem does not need combinatorial optimization, or that the QUBO mapping needs to be improved.   
Moreover, the current construction was inspired by a basic portfolio optimization formulation and neglects possibility of negative $z_i$ and , budget constraints (which would fix the size of $\{r_{selected}\}$), and higher-than-quadratic correlations between the reasons, which can be inserted and ``gadgetized'' into quadratic terms~\cite{babbush2013resource} (adding ancillary variables sparsely connected) while still allowing the use of Ising machines. The advantage of using higher locality solvers than quadratic ones for certain optimization problems has recently been demonstrated \cite{dimitry:pubo_2022}.

\subsubsection{Combinatorial Optimization Solvers Selection}

With QUBO instances being NP-Hard in general, it is expected that the combinatorial optimization solver might take more time to find a quality solution than an acceptable user experience might mandate. The solver used for our baseline results is far from being the optimal choice or from being used in the optimal way, hence we can expect great room for improvement in speed and accuracy of the results in Fig.~\ref{fig:main}. 

To show that enhancements are possible, we proceeded performing a few numerical experiments on a subset of instances of BBH for which we substituted the simulated annealing baseline solver either with a hardware-efficient digital implementation (the \emph{Fujitsu Digital Annealer}~\cite{fujitsu_da}), and with an Adaptive Parallel Tempering (APT) solver from USRA~\cite{Aadit_APT_Adaptive_Parallel_2023}. 

\begin{itemize}
    \item {\bf{Speedup potential}} For the digital annealer experiments, we selected the \emph{Logical Deduction - 7 objects} from BBH as the dataset of interest and sampled 20 questions from it. We consider these instances to be typical, resulting in QUBO problems with an average of 900 variables. We observed consistently a difference of at least an order of magnitude (and often two orders of magnitude) in speed to obtain the same highest quality solution between the baseline method and the specialized hardware. In particular, the time-to-solution given by the Fujitsu Digital Annealer, not considering the latency due to cloud access, were all under 2 seconds, which would enable a turnaround time of the entire CR pipeline under 5 seconds as demonstrated in the relevant technical report ~\cite{icosa_whitepaper}.
    \item {\bf{Accuracy potential}} As clear from Fig.~\ref{fig:main}, the \emph{Formal Fallacies} category of the BBH dataset has proven to be particularly difficult for our CR method. Investigations on these QUBO instances indicated that they were also difficult for the baseline simulated annealing solver, which did not converge to a stable minimum by the allotted time. We then opted to employ the APT solver (which has shown to outperform simulated annealing on spin-glass problems) run on an HPC cluster extensively until convergence to access higher quality solutions. 
    
    Out of the 19 questions we considered for analysis, for 14 questions the answers matched. In 4 questions, the solutions obtained by APT resulted in a correct answer, while the solution obtained by the baseline resulted in an incorrect answer. The opposite is true only for 1 question. Solutions obtained by APT tended to contain a larger number of reasons, with APT picking approximately 49.7 reasons on average while simulated annealing picked only approximately 12.7 reasons. Approximately 86.7$\%$ of reasons picked as part of APT's solution was not part of the reasons picked as part of the simulated annealing solution. This seems to indicate that having a high-quality solution to the QUBO corresponds to final prompts contain a CoT with a substantial number of reasons.
\end{itemize}

It should be noted that simulated annealing, (adaptive) parallel tempering and the digital annealers used in our studies are all driving the optimization through a Monte-Carlo procedure employing thermal relaxation dynamics as the engine. Besides further sophistication of the algorithms~\cite{mohseni2022ising} we could also consider other types of Ising machines, including \emph{quantum} solvers that might access unmatchable sources of speedup, in principle. Based on recent results, particularly promising options in the near-term for time-sensitive applications are superconducting devices implementing quantum annealing~\cite{tasseff2022emerging} as well as gate-model algorithms that exploit the noise as a drive~\cite{maciejewski2024improving}.

\subsubsection{Post-processing of the reasons for the final prompt} 

When reviewing the reasons sampled from the LLM, there were still quite a few reasons that were semantically the same, but identified as different after the semantic matching. This suggests that we could further tune the semantic similarity threshold or potentially leverage an LLM call to help disambiguate reasons. After selecting the ``best" reasons from the combinatorial optimizer, we often find that the prompt does not flow in a logical manner. This is because reasons are often ``consequences" of other reasons and are directional. When the consequences are beyond 2 levels (Reason A causes Reason B which causes Reason C), the combinatorial optimizer doesn't work so well, possibly because it naturally only looks at the correlation of 2 reasons. Futhermore, the output of the combinatorial optimizer is a flat list of reasons with no specific ordering or relationship between the reasons within the prompt. An intermediate step that could be experimented with is to create a CoT or a ``Tree-of-Thought" using an ordered set of reasons. There are recent advances in prompt engineering methods employing similar sophisticated strategies \cite{yao_tree_2023,besta_graph_2024}.

\subsection{Generalizations of the framework}

\subsubsection{Understanding performance characteristics of sampling reasons from an LLM}
CR is based on the assumption that sampling the LLM for reasons will create a distribution of reasons that can be be mapped to an optimization procedure. Exactly how the distribution is created and the accuracy of such distribution needs to be quantified. Factors such as the temperature of the LLM and what type of LLM is being used can impact the distribution of the reasons sampled. The distribution of the reasons are likely to also indicate if the CR procedure is even needed, which could lead to further optimizations.

\subsubsection{Integration with Theorem Provers}
Manual inspection of the reasons that are selected by the combinatorial optimizer reveals that we sometimes find reasons that conflict with each other. We conjecture for harder problems, the number of conflicting reasons will increase and can be removed by leveraging a theorem prover such as \cite{Z3} to further improve the performance of combinatorial reasoning. One of the challenges of using a theorem prover is that they do not scale to thousands of variables. However, using the theorem prover as a post-processing step after the combinatorial optimizer narrows down the reasons to several dozen reasons is practical. The combination of a probabilistic solver combined with a deterministic solver allows for reasoning on open domain problems.

\subsubsection{Integration with Retrieval Augmented Generation}

Retrieval augmented generation (RAG) is used for knowledge intensive tasks. In the real world human problem solving, a combination of knowledge retrieval and reasons is used. A potential avenue for integrating RAG could be:
\begin{itemize}
    \item Use input query to perform a semantic search on a knowledge base to create a knowledge context
    \item Include the knowledge context the prompt when performing sampling of reasons
\end{itemize}
With context windows of 1M tokens \cite{gemini_team_gemini_2024}, the ``reasons" sampled from the LLM could be derived from very long form documents. 

\section*{Acknowledgments}
The work was funded by Icosa Computing Inc. A.A.A., P.A.L. and D.V. were funded by NSF CCF (grant \#1918549) and NASA Academic Mission Services contract NNA16BD14C – funded under SAA2-403506. We thank David Fellah for helpful discussions. We acknowledge the generous access provided by Fujitsu Limited to their Digital Annealer. We thank Michiyuki Tanaka, Yasuyuki Tanaka, Hirofumi Ukita, and Atsushi Kasugai from the Fujitsu team for their support. We also thank Google Deepmind and Google Cloud teams for their comments.

\appendix

\section{Data Availability}
The LLM samples, generated prompts, human analysis and relevant code are available at https://github.com/Icosa-Computing/cr-paper.

\section{End-to-End Example}\label{app:end2end}

Here we provide a full walkthrough of combinatorial reasoning for a particular question from BBH task \textit{snarks}. The original question is presented below:

\begin{mdframed}[style=MyFrame,nobreak=true]
{\fontfamily{qcr}\selectfont\small
Q: Which statement is sarcastic?\\
Options:\\
(A) I'll just bring my keyboard and mouse to the computer and plug it all in\\
(B) I'll just bring my keyboard and mouse to the bus and plug it all in}
\end{mdframed}

\subsubsection{System Instruction}
\begin{mdframed}[style=MyFrame,nobreak=true]
{\fontfamily{qcr}\selectfont\small
Let's think step by step. After each step, condense the reasoning in the step into a sentence and put it in curly braces.
}
\end{mdframed}
\subsubsection{Initial Prompt}\label{app:sampling}
\begin{mdframed}[style=MyFrame,nobreak=true]
    {\fontfamily{qcr}\selectfont\small
Output template:\\
Step 1: reasoning \\
\{condensed reason\}\\
Step 2: reasoning \\
\{condensed reason\} ....\\
Q: Which statement is sarcastic?\\
Options:\\
(A) I'll just bring my keyboard and mouse to the computer and plug it all in\\
(B) I'll just bring my keyboard and mouse to the bus and plug it all in
}
\end{mdframed}

\subsubsection{Reason Extraction}
Each of the $N$ responses will contain a reasoning path, with condensed reasons placed inside curly braces. We extract these condensed reasons and embed them to perform the QUBO mapping. Here, we display two of the LLM responses from the initial sampling phase and demonstrate an example of reason embedding.

\subsubsection{Sampled Response \#1}
\begin{mdframed}[style=MyFrame,nobreak=true]
{\fontfamily{qcr}\selectfont\small
Step 1: Plugging in a keyboard and mouse to a computer is a typical and expected action. \\ \{Plugging in a keyboard and mouse to a computer is a normal occurrence.\}\\Step 2: Plugging in a keyboard and mouse to a bus is an illogical and sarcastic statement. \\\{Plugging in a keyboard and mouse to a bus is absurd and sarcastic.\}\\Conclusion: Option (B) I'll just bring my keyboard and mouse to the bus and plug it all in is sarcastic.\\ \{Option (B) is sarcastic.\}
}
\end{mdframed}

\begin{figure*}[t]
    \centering
    \captionsetup{width=0.75\linewidth}
    \includegraphics[width=1\linewidth]{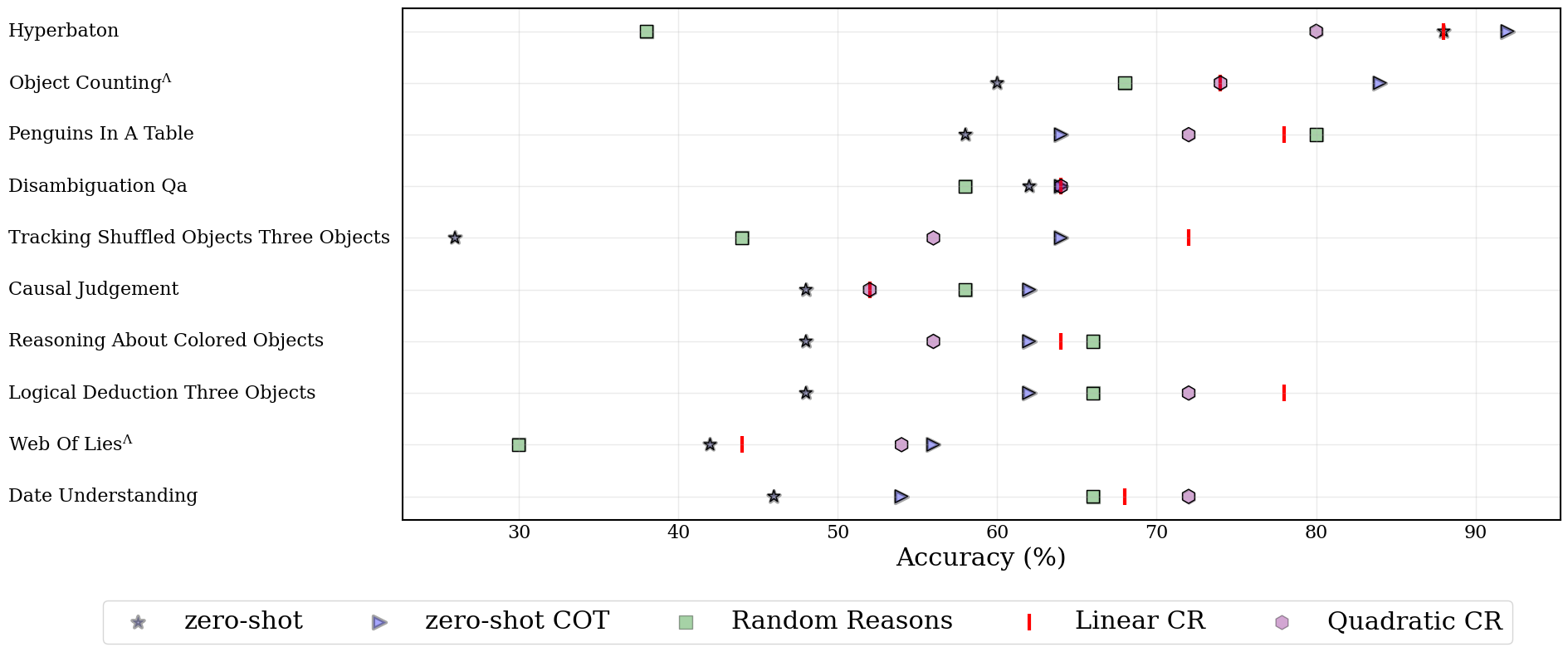}
    \caption{Baseline analysis for Quadratic CR (same as main text) with Linear CR and Random Reasons. Overall performance across the ten datasets were Quadratic CR: $65.2\%$, Linear CR: $68.2\%$, Random: $57.4\%$. 0-shot and 0-shot CoT results are included for reference. The individual tasks are ordered according to the performance of 0-shot CoT.}
    \label{fig:baseline}
    \vspace{3mm}
\end{figure*}

\subsubsection{Sampled Response \#2}
\begin{mdframed}[style=MyFrame,nobreak=true]
{\fontfamily{qcr}\selectfont\small
Step 1: Plugging a keyboard and mouse into a computer is a common and expected action, so the statement using this scenario is not sarcastic.\\
\{Plugging a keyboard and mouse into a computer is a common and expected action.\} \\
Step 2: However, plugging a keyboard and mouse into a bus doesn't make sense as buses do not have such connections, making this statement sarcastic. 
\{Plugging a keyboard and mouse into a bus is not a logical or realistic scenario, indicating sarcasm.\}

The sarcastic statement is:
(B) I'll just bring my keyboard and mouse to the bus and plug it all in.
}
\end{mdframed}
We focus on the following two reasons appearing in the two different LLM outputs (only curly brackets are extracted).
\begin{itemize}
{\fontfamily{qcr}\selectfont\small
\item Plugging a keyboard and mouse into a computer is a common and expected action.
\item Plugging a keyboard and mouse into a computer is a normal occurence.
}
\end{itemize}
The Sentence Transformer embeds them into vectors $\bf{\gamma}_1$ and $\bf{\gamma}_2$. They are semantically identical, as determined by their inner product $\bf{\gamma}_1 \cdot \bf{\gamma}_2 = 0.95$, which is well above the threshold of 0.90. These reasons will be collapsed into one single distinct reason in the CR pipeline.

\subsubsection{Ising Mapping and Solving}

The extracted reasons will be mapped to a quadratic optimization problem according to the procedure outlined in \ref{qubo_mapping}. For this example question, the QUBO instance contains $174$ variables. The best solution found by simulated annealing has an energy of $-806.3$. We map the best solution back to a set of 45 selected reasons and corresponding $w$-$value$s (which end up being only two different values). 

\subsubsection{Final Prompt} \label{app:finalPrompt}

Using the selected reasons and their w-values, we construct the final prompt. The LLM is queried with the final prompt and system instructions at temperature $T = 0$. We have omitted the mid-section of the prompt due to space constraints. The full prompt for the example and all other prompts can be found in the Supplementary Material.

\subsubsection{System Instruction}\label{sec:sys_instruc}
\begin{mdframed}[style=MyFrame,nobreak=true]
{\fontfamily{qcr}\selectfont\small
Each W-Statement starts with the substring (w), where (w) is a number called the W-Value. Identify and state each W-Value. W-Statements with higher W-Values have more reliable information. You may not provide multiple possible answers, you must narrow your final solution down to a single answer. Refer to each W-Statement and their W-Values in your reasoning. Your final answer must be of the form SOLUTION: (option).
}
\end{mdframed}
\subsubsection{Prompt}
\begin{mdframed}[style=MyFrame,nobreak=False]
{\fontfamily{qcr}\selectfont\small
Q: Which statement is sarcastic?\\
Options:\\
(A) I'll just bring my keyboard and mouse to the computer and plug it all in\\
(B) I'll just bring my keyboard and mouse to the bus and plug it all in\\
W-Statements:\\
(0.028) Using a keyboard and mouse on a bus is an absurd idea.\\
(0.028) The statement "I'll just bring my keyboard and mouse to the bus and plug it in" is sarcastic because it suggests an action that is illogical.\\
...\\
(0.009) Option (B) is sarcastic as it involves an absurd suggestion.\\
(0.009) Option (B) is sarcastic as it contains an unrealistic scenario, making it an unlikely action.\\
(0.009) Option (B) "I'll just bring my keyboard and mouse to the bus and plug it all in" is sarcastic.\\
(0.009) B is the sarcastic statement\\
}

\end{mdframed}
\subsubsection{LLM Response}
\begin{mdframed}[style=MyFrame,nobreak=False]
{\fontfamily{qcr}\selectfont\small
The W-Value for all statements is 0.028 except for the last two, which have a W-Value of 0.009. The majority of the statements with a W-Value of 0.028 clearly identify option (B) as the sarcastic statement, while the statements with a W-Value of 0.009 do not provide a definitive answer. Therefore, the final solution is:\\\\
SOLUTION: (B) I'll just bring my keyboard and mouse to the bus and plug it all in
}
\end{mdframed}

\section{Evaluation of QUBO mapping and solving }\label{append:QUBOs}
In Figure \ref{fig:baseline}, we investigated the performance differences between CR as per main text (``Quadratic CR'') versus selecting a set $\{r_{selected}\}$ of reasons either randomly (``Random Reasons'') among all the $\{r_{distinct}\}$ possible or the set that optimize the cost function $\sum_i^{\{r_{selected}\}} \tilde{l}_i(\mu,m_i,\alpha,\kappa,x_i)$ from Eq. \ref{eq:linear_terms} (``Linear CR''). We restrict the analysis to the top 10 categories with the minimum value of the ratio between the cardinality of $\{r_{selected}\}$ and $\{r_{distinct}\}$, which we take as a proxy for ``reasoning difficulty'' (see Table \ref{tab:quadratic percentages}). Results indicate that random selection is generally performing worse and that quadratic terms seem to be useful only for 2/10 categories, which happen to be also among the most difficult (in terms of accuracy) for zero shot methods. We believe this correlation between difficulty and importance of quadratic terms in the QUBO deserves more investigation.

We performed a human evaluation in order to understand the QUBO model effect on correlated reasons to verify that $c_{ij}$ and $q_{ij}$ could be seen as measures of strength of relationship between two sampled reasons. Restricting our analysis to the \emph{causal judgement}, \emph{movie recommendation}, and \emph{sports understanding} datasets, we assigned a value (-1,0,+1) to each pair of reasons appearing in 10 samples depending on whether the pair appears respectively inconsistent, consistent or independent in the eyes of a human. A first observation is that, \emph{within each sample}, reasons that are inconsistent are very rare (they appeared only in one sample), which reassure that the LLM does not output contradictory answers for the dataset examined. Across samples, in \emph{causal judgement}, we observed that the 50 pairs with higher values of $c_{ij}$ and $q_{ij}$ did contain a large fraction of correlated and consistent pairs. However, on other datasets such as \emph{movie recommendation} or \emph{sports judgement}, there were a significant fraction of correlated and consistent reasons even in the lowest 50 pairs. Moreover, many of the reason pairs with the lowest 50 values seem semantically identical - indicating a failure of the similarity measure described in \ref{sec:sampling}. Based on this and other anecdotal evidence, we hypothesize that for easy questions, the LLM might be producing correlated reasons consistent with the correct answer nearly all the time. Results call to look at the pairs resulting in negative $c_{ij}$ much more closely in future studies.

We performed human evaluation of the consistency of the final reasons selected after the QUBO Solver $\{r_{selected}\}$ for the same restricted dataset, observing that inconsistency is very rare (for \emph{casual judgment}) or difficult to ascertain for the \emph{movie recommendation}, and \emph{sports understanding} datasets. A preliminary observation indicates for NLP tasks we match or slightly outperform the simple zero-shot CoT method (without human-in-the-loop), while for algorithmic tasks CR usually has slightly lower accuracy.

\bibliographystyle{kr}
\bibliography{kr-sample}

\end{document}